\documentclass[sigconf,natbib=true,anonymous=false]{acmart}

\usepackage{pifont}
\newcommand{\cmark}{\ding{51}}%
\usepackage[T1]{fontenc}
\newcommand{\xmark}{\ding{55}}%
\usepackage{makecell}
\usepackage{cellspace}
\usepackage{subfigure}
\usepackage{multirow}
\usepackage{balance}
\usepackage{graphicx}
\usepackage{soul} 
\usepackage{color, xcolor} 
\definecolor{LightSkyBlue}{RGB}{179,247,247}
\definecolor{mygreen}{HTML}{cce5f7}
\definecolor{myred}{HTML}{F19C99}

\pdfoutput=1

\AtBeginDocument{%
  \providecommand\BibTeX{{%
    \normalfont B\kern-0.5em{\scshape i\kern-0.25em b}\kern-0.8em\TeX}}}

\setcopyright{acmcopyright}
\copyrightyear{2023}
\acmYear{2023}
\acmDOI{XXXXXXX.XXXXXXX}

\acmConference[SIGIR '23]{The 46th International ACM SIGIR Conference on Research and Development in Information Retrieval}{July 23--27, 2023}{Hybrid Conference,Taipei, Taiwan}
\acmBooktitle{The 46th International ACM SIGIR Conference on Research and Development in Information Retrieval (SIGIR '23), July 23--27, 2023, Hybrid Conference, Taipei, Taiwan}
\acmPrice{15.00}
\acmISBN{978-1-4503-XXXX-X/18/06}

\acmSubmissionID{5687}

\begin{document}

\title[MoocRadar: A Fine-grained and Multi-aspect Knowledge Repository \\ for Improving Cognitive Student Modeling in MOOCs]{MoocRadar: A Fine-grained and Multi-aspect Knowledge Repository for Improving Cognitive Student Modeling in MOOCs}



\author{Jifan Yu}
\affiliation{%
  \institution{DCST, Tsinghua Univerisity}
  \city{Beijing 100084}
  \country{China}
}
\email{yujf21@mails.tsinghua.edu.cn}

\author{Mengying Lu}
\affiliation{%
  \institution{SIGS, Tsinghua Univerisity}
  \city{Shenzhen 518055}
  \country{China}
}
\email{lumy22@mails.tsinghua.edu.cn}

\author{Qingyang Zhong}
\affiliation{%
  \institution{DCST, Tsinghua Univerisity}
  \city{Beijing 100084}
  \country{China}
}
\email{zqy20@mails.tsinghua.edu.cn}

\author{Zijun Yao}
\affiliation{%
  \institution{DCST, Tsinghua Univerisity}
  \city{Beijing 100084}
  \country{China}
}
\email{yaozj20@mails.tsinghua.edu.cn}

\author{Shangqing Tu}
\affiliation{%
  \institution{DCST, Tsinghua Univerisity}
  \city{Beijing 100084}
  \country{China}
}
\email{tsq22@mails.tsinghua.edu.cn}

\author{Zhengshan Liao}
\affiliation{%
  \institution{IoE, Tsinghua Univerisity}
  \city{Beijing 100084}
  \country{China}
}
\email{liaozs20@mails.tsinghua.edu.cn}

\author{Xiaoya Li}
\affiliation{%
  \institution{IoE, Tsinghua Univerisity}
  \city{Beijing 100084}
  \country{China}
}
\email{lixiaoya21@mails.tsinghua.edu.cn}

\author{Manli Li}
\affiliation{%
  \institution{IoE, Tsinghua Univerisity}
  \city{Beijing 100084}
  \country{China}
}
\email{marylee@mail.tsinghua.edu.cn}

\author{Lei Hou}
\affiliation{%
  \institution{BNRist, DCST, Tsinghua Univerisity}
  \city{Beijing 100084}
  \country{China}
}
\email{houlei@tsinghua.edu.cn}

\author{Hai-Tao Zheng}
\email{zheng.haitao@sz.tsinghua.edu.cn}
\affiliation{%
  \institution{SIGS, Tsinghua University}
  \institution{\& Pengcheng Laboratory}
  \city{Shenzhen 518055}
  \country{China}
}

\author{Juanzi Li}
\authornote{Corresponding author.}
\affiliation{%
  \institution{BNRist, DCST, Tsinghua Univerisity}
  \city{Beijing 100084}
  \country{China}
}
\email{lijuanzi@tsinghua.edu.cn}

\author{Jie Tang}
\affiliation{%
  \institution{BNRist, DCST, Tsinghua Univerisity}
  \city{Beijing 100084}
  \country{China}
}
\email{jietang@tsinghua.edu.cn}

\renewcommand{\shortauthors}{Yu, et al.}

\begin{abstract}
    Student modeling, the task of inferring a student's learning characteristics through their interactions with coursework, is a fundamental issue in intelligent education. Although the recent attempts from knowledge tracing and cognitive diagnosis propose several promising directions for improving the usability and effectiveness of current models, the existing public datasets are still insufficient to meet the need for these potential solutions due to their ignorance of complete exercising contexts, fine-grained concepts, and cognitive labels. In this paper, we present MoocRadar, a fine-grained, multi-aspect knowledge repository consisting of $2,513$ exercise questions, $5,600$ knowledge concepts, and over $12$ million behavioral records. Specifically, we propose a framework to guarantee a high-quality and comprehensive annotation of fine-grained concepts and cognitive labels. The statistical and experimental results indicate that our dataset provides the basis for the future improvements of existing methods. Moreover, to support the convenient usage for researchers, we release a set of tools for data querying, model adaption, and even the extension of our repository, which are now available at \url{https://github.com/THU-KEG/MOOC-Radar}.
    
\end{abstract}




\maketitle

\begin{figure*}
    \centering
    \includegraphics[width=0.9\linewidth]{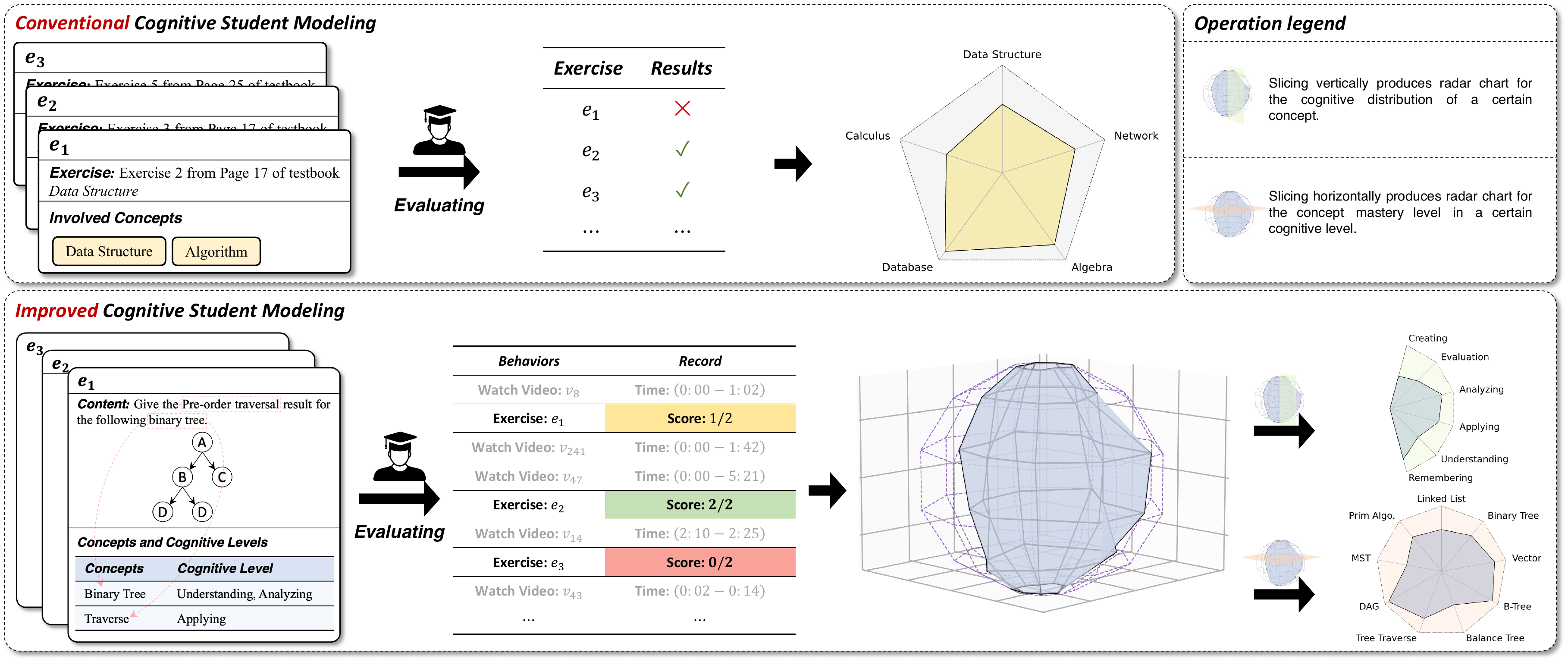}
    \caption{An illustration of the cognitive student modeling topic. Based on the the knowledge tracing or cognitive diagnosis methods, MoocRadar provides a basis for extending this task to a more informative setting with exercise contents, fine-grained concepts, cognitive labels and other learning behaviors.}
    \label{fig:demon}
\end{figure*}

\section{Introduction}


Since the first proposal of the idea of \emph{Intelligent Tutoring System}~\cite{psotka1988intelligent}, constructing the student model~\cite{vanlehn1988student} that can infer the characteristics of each individual student has been a fundamental stage for supporting AI-driven educational services, such as study planning~\cite{xia2019peerlens,pan2017prerequisite}, learning resource recommendation~\cite{wan2019hybrid,bulathwela2020truelearn} and teaching assistant chatbot~\cite{zhu2021proactive}. Compared with modeling the explicit features of a student (\emph{e.g.}, \emph{Age}, \emph{Grade})~\cite{feng2019understanding}, modeling the implicit and cognitive features, such as approximating a student's proficiency level on specific knowledge~\cite{corbett1994knowledge}, is more emergent and challenging, which attracts abundant research interests on the relevant topics like \emph{Knowledge Tracing}~\cite{piech2015deep,shen2020convolutional} or \emph{Cognitive Diagnosis}~\cite{wang2020neural}.

Figure \ref{fig:demon} shows an illustration of the necessary elements for completing such tasks: Given the students' behavior records (\emph{e.g.}, right or wrong) on a series of exercise questions, the goal is to estimate their actual knowledge state on the corresponding concepts (\emph{e.g.}, \emph{Database}). To achieve an applicable performance, except for the early attempts that simply employ a neural method to model the above elements within a sequence architecture~\cite{piech2015deep,zhang2017dynamic}, researchers pour great efforts into seeking the beneficial side information~\cite{shen2021learning}, mining appropriate knowledge associations~\cite{abdelrahman2019knowledge,lee2022contrastive} and even stimulating the students' cognitive process~\cite{shen2021learning} for lifting the accuracy and interpretability of the proposed methods~\cite{wang2020neural}.


However, due to the limited access to data resources and the costly expert annotation, there remain several inherent defects of the existing popular datasets~\cite{feng2009addressing,choi2020ednet} that may hinder these promising explorations, which can be summarized as three major points:

$\bullet$ \textbf{Incomplete Context of Exercising Behavior}: Although diverse, relevant contexts of students' exercising behaviors, \emph{e.g.}, the instructional structure and hierarchy of exercises~\cite{pandey2020rkt}, the exercise content and types~\cite{liu2019ekt}, and associated learning behaviors~\cite{mao2021learning}, have been proven to be beneficial to the modeling, current datasets~\cite{choi2020ednet} usually consider few types of them, which cannot support more feature combinations for utilizing these features.

$\bullet$ \textbf{Coarse Granularity of Knowledge Concept}: Another common issue is that most of the existing datasets are only annotated with coarse-grained concepts, \emph{e.g.}, there are only $123$ concepts for the $26,688$ exercise questions in ASSISTment2009~\cite{feng2009addressing}, while only $41$ concepts are labeled for $722$ problems in Junyi~\cite{JunyiOnlineLearningDataset}. Although there are some data resources that provide fine-grained concepts~\cite{yu2021mooccubex}, they mostly lack expert annotation~\cite{li2019should} and ignore the scenarios that one exercise can match multiple concepts.

$\bullet$ \textbf{Neglected Annotation of Cognitive Process}: Despite the growing attention to modeling student's cognitive process, such as the learning and forgetting regulation~\cite{shen2021learning}, question difficulty levels~\cite{wang2020neural}, building the supervision labels from the view of cognitive pedagogy is still an insufficient topic for educational dataset construction~\cite{phan2010students}. How to appropriately invoke such guidance for modeling cognitive processes remains a challenging topic.

In this paper, we propose MoocRadar, a large knowledge repository for cognitive student modeling. Built upon open data sources~\cite{yu2021mooccubex} and abundant expert annotation, MoocRadar preserves $2,513$ exercise questions with their multi-aspect contexts, $5,600$ fine-grained concepts, $14,224$ online learning students and $12,715,126$ behavior records. Specifically, all the learning behavior is labeled by educational experts for building cognitive-aware supervision, following the widely accepted \emph{Bloom Cogntive Taxonomy}~\cite{krathwohl2002revision}, which provides a brand new view for future developments of relevant topics. To host the whole construction, we design a framework that consists of heterogeneous data integration, automatic concept recognition, and a series of mechanisms for convenient annotation and quality control. Furthermore, we develop and publish an affiliated toolkit for multiple common queries and representation training, which may help researchers efficiently utilize and apply our resources to improve their models and systems.

We conduct a series of statistical and experimental investigations to present the characteristics of MoocRadar and whether sufficient contexts, fine-grained concepts, and cognitive labels indeed benefit current methods. Moreover, based on our released repository, it is feasible to extend the current \emph{Knowledge Tracing} or \emph{Cognitive Diagnosis} tasks to a deeper and more informative setting as shown in Figure \ref{fig:demon}, which can provide a more detailed demonstration for presenting the fine-grained and multi-aspect cognitive states of a certain student. We hope that our efforts can call for research interests to exploit pedagogical and cognitive theories and build more intelligent next-generation educational applications.
 
\textbf{Contributions}. Our contributions can be summarized as: 

(1) An open-access, fine-grained, and multi-aspect knowledgeable repository with high-quality annotation for serving and promoting the research of cognitive student modeling. 

(2) A framework that consists of a series of standards and a toolkit for convenient usage and extension on educational tasks. 

(3) An investigation and several primary insights about how to invoke pedagogical theories for refining the student modeling task and achieving more applicable services of AI-driven education.






\begin{figure*}
    \centering
    \includegraphics[width=0.98\linewidth]{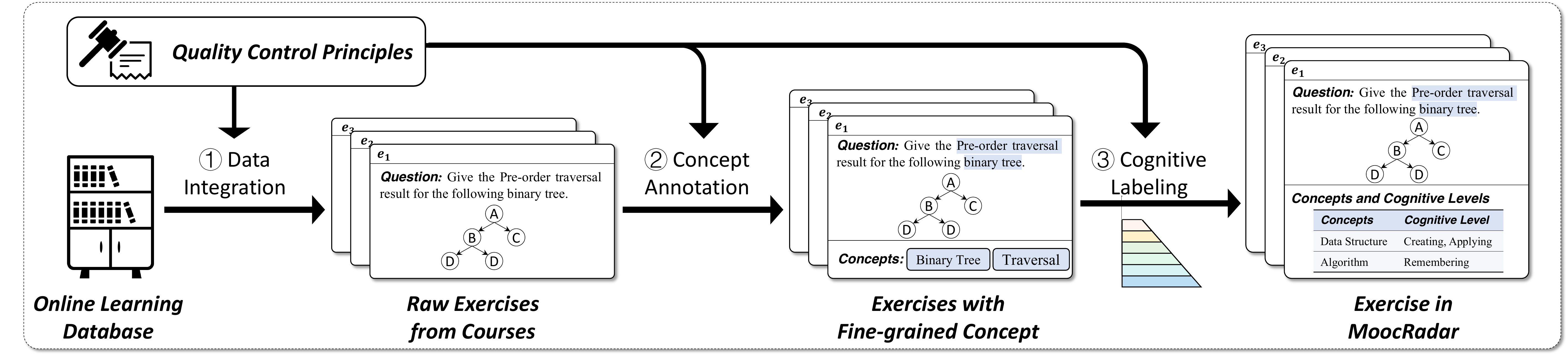}
    \caption{The overall construction framework of MoocRadar from the view of exercises. With the refinement of the exercises, the heterogeneous information is collected correspondingly. All the processing operations are under strict quality control.}
    \label{fig:framework}
\end{figure*}

\section{Background}

\subsection{Cognitive Student Modeling}

Student Modeling (or the \emph{Student Model}) is an indispensable and basic concept of a variety of educational technology theories~\cite{vanlehn1988student}, whether the early \emph{Intelligent Tutoring System}~\cite{anderson1985intelligent} or the recent \emph{Adaptive Learning}~\cite{paramythis2003adaptive}. This essential task actually covers the modeling of students' various learning characteristics, including the explicit features (\emph{i.e.}, basic information like \emph{Gender}, \emph{Age}) and the implicit profiles (\emph{e.g.}, knowledge cognitive state and learning style)~\cite{chrysafiadi2013student}, which is a prior stage for downstream applications such as educational recommendation~\cite{bulathwela2020truelearn} and instructional games~\cite{abdelrahman2019knowledge}.

With the accumulation of abundant learning behavioral data from massive open online courses (MOOCs)~\cite{yu2020mooccube}, inferring the student's cognitive states of specific knowledge gradually becomes a promising topic~\cite{piech2015deep}. Despite the different research origins, contributors in both \emph{Knowledge Tracing}~\cite{liu2019ekt} and \emph{Cognitive Diagnosis}~\cite{wang2020neural} are currently concentrated on predicting students' future performance through mining their exercise records, which can be easily adapt to the cognitive student modeling task.

Empowered by the technical advances such as sequence to sequence learning~\cite{abdelrahman2019knowledge,wang2020kerl} and contrastive self-training~\cite{lee2022contrastive}, there are several growing directions in improving such cognitive student modeling: (1) \emph{Enriching the Information of Exercises}: Another intuitive direction is to exploit the rich side information in modeling learning resources, such as the textual contents and structures of the exercises~\cite{pandey2020rkt,liu2019ekt}, as well as the other behavioral records from relevant students~\cite{shen2020convolutional};  (2) \emph{Refining the Modeling of Knowledge Concepts}: Mining the deep meaning and inner relationships of the concepts behind the exercises is one of the primary directions, \emph{e.g.}, \citet{chen2018prerequisite} employ the prerequisite relations of concepts and \citet{tong2020structure} design an independent module for modeling concepts; (3) \emph{Stimulating the Cognitive Process of Learning}: There is another rising trend of invoking the pedagogical guidance and stimulating the students' cognitive learning process, such as learning and forgetting regulation~\cite{chen2022knowledge}, question difficulty levels~\cite{wang2020neural}, for explaining the reason of corresponding student performance. These directions can also be combined~\cite{shen2021learning} and deployed in adaptive learning~\cite{zhong2022towards}, which continuously triggers the discussions about employing more features to enhance the current methods~\cite{abdelrahman2022knowledge}.

\subsection{Open Educational Datasets}

Compared with the prosperous improvement of student modeling methods, the upgrading, and proposal of open-source educational datasets are respectively limited~\cite{liu2021survey,abdelrahman2022knowledge,zhao2022edukg}. In general, the source of datasets can be roughly divided into two categories: \emph{Public AI-driven Education Challenge} and \emph{Open-access Education Repository}.

The \emph{Public AI-driven Education Challenge} is a natural and easy-to-use resource for researchers, which is usually hosted by a particular online learning platform or group, such as KDD cups (Algebra 2005-2006, Bridge 2006-2007)~\cite{stamper2010challenge}, ASSISTment~\cite{feng2009addressing,pardos2013affective} and Junyi Academy~\cite{JunyiOnlineLearningDataset}. Despite their convenience and popularity, they make concessions in data coverage for graceful task formalization. For all types of relevant behavioral and structured data, most of them only provide the basic identifiers (\emph{e.g.}, the concepts in ASSISTment are only concept ids), not the actual and complete content.

The \emph{Open-access Education Repositories} are more inclined to provide useful information from multiple perspectives, which are mainly established by integrating processed data from heterogeneous sources, such as STATIC2011~\cite{koedinger2010data} EdNet~\cite{choi2020ednet}, and MOOCCubeX~\cite{yu2021mooccubex}. Due to their giant sizes, they usually reduce the annotation scale to control the cost, which results in a lack of fine-grained knowledge concepts. Even though some previous efforts~\cite{yu2021mooccubex} attempt to extract concepts from video texts with weak supervision automatically, the high-quality concept annotation of exercises is still ignored and fairly expensive.

Meanwhile, because of the absence of cognition-aware labels, current methods that follow learning regulations~\cite{chen2022knowledge} can only build pseudo labels according to other behavioral records. Therefore, we feel obliged and emergent to conduct adequate expert annotation and propose MoocRadar in order to support relevant researchers.

\section{MOOC Radar}

In this section, we first propose the principles and the overall framework of the construction of MoocRadar, then present the detailed workflow about how to conduct the large-scale and high-quality concept \& cognition annotation effectively. Finally, we introduce access to our repository and the affiliated toolkit.

\textbf{Principles.} To guarantee the usability of the proposed data resources, we identify that there are two major principles when constructing the repository. (1) \emph{Exercise-centered Data Organization}: As exercising is the most crucial behavior in current student modeling methods~\cite{piech2015deep}, it is appropriate to prioritize access to students' exercising sequences and then retrieve diverse and related information based on these indexes; (2) \emph{Comprehensive Expert Annotation}:
Since knowledge concepts and cognitive levels are highly specialized content, we must have a sufficient number of experts for manual annotation, even if this matter has been commonly replaced by automatic means in previous efforts due to its high cost~\cite{yu2021mooccubex}.

\textbf{Framework.} As illustrated in Figure \ref{fig:framework}, starting from the large-scale collection and processing of online learning data, we host a series of expert annotations to label the fine-grained concepts and cognitive levels of the exercises with proper intelligent assistance. After quality inspection, the remaining data is further stored and packaged to be easily accessed as an informative resource. This three-module framework can be decomposed as follows:

(1) \emph{Heterogeneous Data Integration}: Given the semi-structured and weakly supervised online learning data, this stage mainly select the high-quality exercising records and collect the heterogeneous related contexts, which builds the data basis of the whole repository.

(2) \emph{Fine-grained Concept Annotation}: In this stage, we conduct the annotation for fine-grained concepts. Specifically, we propose a pre-trained model based concept acquisition method for reducing the workload and assisting the annotators.

(3) \emph{Cognitive Exercise Labeling}: Following the Bloom Cognitive Taxonomy~\cite{krathwohl2002revision}, we first conduct text analysis for the preparation of cognition-aware labeling, and then complete the large-scale expert annotation. Finally, all the annotation results are aligned with the concept labels and packaged with strict quality control.

\subsection{Heterogeneous Data Integration}

Our data is from MOOCCubeX~\cite{yu2021mooccubex}, a large-scale and knowledge-centered online education repository that contains $358,265$ exercises, $637,572$ automatically extracted concepts (with noises), and $296$ million of student behavioral data from $4,216$ courses. As an open data resource upon XuetangX\footnote{\url{https://www.xuetangx.com/}}, one of the largest MOOC platforms in China, it also provides abundant, well-maintained learning material and behavioral records with rich relationships. Therefore, we select it as our original data resource and initiate subsequent data integration processes.

\subsubsection{Course Selection} 

Through a pioneer data observation, we discover that a critical premise to ensure the informativeness of the exercises, the richness of the features, and the adequacy of the associated learning behaviors is the proper selection of corresponding courses. For those courses marked as ``Outstanding'' by the platform, they have sufficient content and reasonable exercises and therefore attract a high volume of students. Therefore, we select a total of $154$ ``Outstanding'' courses from $12$ fields, such as engineering, law, medicine, and economics, according to the quantitative distribution of various subjects. The average enrollment of these courses, each from a different instructor, is $68,941$, which is ample to support the following data collection and labeling.

\subsubsection{Data Collection} 

Based on the selected course, we need to collect multi-aspect data about the exercises. 

(1) Exercise Content and Answer: The content of an exercise is proven to be useful in student modeling~\cite{liu2019ekt}, which can also support the tasks like exercise retrieval. We obtain all the textual content and the types (\emph{e.g.}, single-choice questions, multiple-choice questions) of each exercise problem. Furthermore, the standard answer of exercises is often ignored in previous studies, which we also collect as additional side information in this stage.

(2) Exercise Graph: Another beneficial information is the structural relations of the exercises~\cite{pandey2020rkt}. As the exercises are orderly organized in the chapters of the courses, we employ these natural structures to build the exercise graph in our dataset, which can be summarized as a $3$-level hierarchy and a partial order graph.  

(3) Exercising Behaviors: Once the exercises are determined, we can crawl the student behavioral records on these problems. For each student, we sort their exercise records in time order and collect the timestamp to build an interaction sequence. Specifically, thanks to the collection of the standard answer, we can automatically assign the performance score of each behavior. We further consider the \emph{Half Right} scene that a student gives a partially correct answer in a multiple choice question and classify the scores into three levels: \texttt{0}-Wrong, \texttt{1}-Half Right, \texttt{2}-Right. Researchers can decide whether to reduce it to a dichotomy when employing our dataset.

(4) Associated Resources and Behaviors: Except for the exercising information and behaviors, there are other types of relevant features, such as the course videos that are in the same chapters, video-watching behaviors, and the course descriptions. We crawl all this information for future exploitation in improving student modeling as well as other relevant topics such as adaptive learning recommendation~\cite{liu2019exploiting}. Furthermore, as MOOCCubeX provides the concept set of these resources, we preserve them as a distant supervision dictionary for subsequent annotation.

\subsubsection{Noise Filtering}

To mitigate training problems caused by data quality, such as overfitting, model traversal, and failure to converge, we performed basic filtering on the constructed student behavior sequences. First, we remove behavioral sequences that are too short as well as score anomalies (\emph{e.g.}, all-correct sequences based on potential cheating means) based on the distribution of the existing datasets~\cite{feng2009addressing,JunyiOnlineLearningDataset}. Second, we filter out the exercises that the contents are incomplete. Once a course cannot preserve enough exercises, the whole courses with its all exercises are excluded. Finally, we preserve only $40\%$  of the original data to be annotated.

\subsection{Fine-grained Concept Annotation}
\label{sec:concept_annotation}

Concepts, \emph{i.e.}, the knowledge concepts implied after the exercise content (such as \emph{Graph Neural Network} in the Artificial Intelligence course), which are highly essential and specialized components of the knowledge tracing tasks. Even for the professional educators in the corresponding courses, it is non-trivial to annotate accurate fine-grained concepts from scratch. Given the automatically extracted concepts from MOOCCubeX, we first conduct a noise-tolerant concept pre-recognition via large language model prompting, and then conduct extensive annotation with a $40$-people group of experts.

\subsubsection{Discipline-aware Concept Extraction}

The previous concept acquisition mainly relies on traditional \emph{Phrase Mining} or \emph{Entity Linking} solutions. Despite a few efforts that regard this task as \emph{Distantly Supervised Named Entity Recognition (Distant NER)}, there are several remaining issues to be considered. (1) Discipline Variation: For the courses in different disciplines, concepts have large gaps in vocabulary length, style, and semantics, and it is often tricky for existing methods to maintain consistent performance across disciplines. (2) Noisy Dictionary: As the prepared distant supervision is determined to be noisy from the given dictionary, most of the current methods may fail to distinguish the suitable and misleading labels, which eventually leads to entirely unusable extraction results. Therefore, we attempt to exploit the large language model prompting to enhance current distant NER methods with discipline information, which can be further explored in future work. In general, our automatic concept preparation is in two steps.

\begin{table*}[ht]
\centering
\caption{A demonstration of student exercise behavior in MoocRadar. The example shows the exercise behavior of student \emph{U}\_309 in course \emph{Introduction to Economics} and \emph{Machine Learning}. Blue phrases are the annotated multi-level concepts. }
\label{tab:dataset_behavior}
\setlength{\abovecaptionskip}{0.2cm}
\setlength{\belowcaptionskip}{0.cm} 
\setlength{\tabcolsep}{1.2mm}
\begin{tabular}[\linewidth]{|c|c|c|c|c|c|c|c|c|c|}
\toprule
 UserId & Course   & ChapterId & ExerciseId  & Content & Option & Cognitive  & Answer  & Score  &  Time \\\hline
 \multirow{2}{*}{\makecell*[c]{\ \\ \ \\ \ \\ \ \\ \ \\\emph{U}\_309}} &
\multirow{2}{*}{\makecell*[c]{\ \\  \emph{Introduction}\\\emph{to}\\\emph{Economics}}} & \emph{Ch}\_6145 & \emph{Ex}\_12138  &\makecell*[l]{The research objects \\of \sethlcolor{mygreen}\hl{\emph{Microeconomics}}\\ include: ()} &
\makecell*[l]{A:`Single Customer', \\B: `Single Producer', \\C: `Single Market', \\D: `\sethlcolor{mygreen}\hl{\emph{Price Theory}}'\\} 
 &  \makecell*[l]{2: Under-\\standing}  & \makecell*[l]{Ref: ABC\\User: AC} & \makecell*[c]{$1$}  &  \makecell*[l]{2020-\\07-18\\ 15:03:48}\\
\cline{3-10}
&  &  \emph{Ch}\_6145 & \emph{Ex}\_12145&  \makecell*[l]{The starting point \\of \sethlcolor{mygreen}\hl{\emph{Western Economic}}\\ research is ()} & \makecell*[l]{A:`\sethlcolor{mygreen}\hl{\emph{Rareness}}', \\B: `Desire', \\C: `Demand', \\D: `Supply'\\}  &  \makecell*[l]{1: Remem-\\bering} & \makecell*[l]{Ref: A\\User: C} & \makecell*[c]{$0$} & \makecell*[l]{2020-\\07-18\\ 15:04:11} \\ 
\cline{2-10}
& ... & ... & ... & ... & ...  &  ...  & ... & ... & ... \\ 
\cline{2-10}
&  \makecell*[c]{ \sethlcolor{mygreen}\hl{\emph{Machine}}\\\sethlcolor{mygreen}\hl{\emph{Learning}}}  &  \emph{Ch}\_1129 & \emph{Ex}\_26580&  \makecell*[l]{When $k$ is different,\\ the test results of \\  \sethlcolor{mygreen}\hl{\emph{KNN algorithm}}\\ may be the same.} & \makecell*[l]{A: `True', \\ B: `False'\\}  &  \makecell*[l]{3: Applying} & \makecell*[l]{Ref: A\\User: A} & \makecell*[c]{$2$} & \makecell*[l]{2020-\\09-07\\ 17:46:46} \\
\bottomrule
\end{tabular}
\end{table*}

\textbf{Discipline-aware Dictionary Cleaning}. Before the training of distant NER methods, we conduct a preceding step to denoise the automatically extracted concept as distant supervision signal. Specifically, we conduct a discipline classification process via prompt-based learning~\cite{gu2022ppt}. Taking the input of each concept $c_i$ in the provided concept dictionary $T=\{ c_i \} _ {i=1,...,m}$ of MOOCCubeX, the classification returns a ranked list of related disciplines $F_{c_i} \subset F$ and outputs $p_j$ for $ f_j \in F = \{{f_j}\}_{j=1,...,k}$ to indicate its likelihood to be related to $f_j$ discipline, formally:
\begin{equation}
p_j(x^{'}) = LM(f_{fill}(x^{'},f_j);\theta)
\label{eq:distribution}
\end{equation}
where $x^{'} = f_{prompt}(c_i)$ is a prompt with the concept $c_i$ filled template slot $[concept]$,
and function $f_{fill}(x^{'},f_j)$ fills in the slot $[MASK]$ with the potential answer $f_j$.

After that, we can infer the top-$k$ disciplines of a certain concept, which help us exclude the mismatched concepts. We have $\{(X_m,D_m)\}_{m=1}^{M}$ as distantly supervised data, where $X_m=[x_1,\allowbreak x_2,\ldots,x_N]$, composed of $N$ tokens, $D_m=[d_1,d_2,\ldots,d_N]$, based on the BIO schema for the training of concept extraction model.


\textbf{Noise-tolerant Model Self-Training}. We deploy a BERT~\cite{kenton2019bert} model as the backbone to train a fine-grained concept extraction tool. As the common formulation, we use $f(\cdot;\theta)$ to denote our model parameterized by $\theta$, which is a token-wise classifier.
$f_{n,c}(\cdot;\cdot)$ is the probability of the $n$-th token in $X_m$ belonging to the $c$-th class from the BIO schema.
And the model is trained by minimizing the cross entropy loss $\mathcal{L} (\theta)$ over $\{(X_m,D_m)\}_{m=1}^{M}$:
\begin{equation}
\mathcal{L} (\theta) =  \frac{1}{M}\frac{1}{N}\sum _{m=1}^{M}\sum _{n=1}^{N}-\log f_{n,d_{m,n}}(X_m;\theta)
\label{loss}
\end{equation}

To make the model noise-tolerant, we conduct a P/U Learning based self-training. We apply the binary label assignment mechanism for using this algorithm by mapping ``O'' to 0 and ``B'', ``I'' to 1. Then we can get a positive set and unlabeled set to host a new P/U training loss $\widehat{\mathcal{L} } (\theta)$ as \citet{peng2019distantly}.

Therefore, the parameters of our model $\theta^{*}$ are learned by the combination of the cross entropy loss $\mathcal{L} (\theta)$ and the PUL loss $\widehat{\mathcal{L} } (\theta)$:
\begin{equation}
	\theta^{*}=\mathop{argmin}\limits_{\theta} (\mathcal{L} (\theta) + \beta \cdot \widehat{\mathcal{L} } (\theta) )
	\label{pul}
\end{equation}
where a parameter $\beta$ is introduced to balance these two loss functions. Employing this model, we produce a better extractor that recognizes candidate concepts for each exercises, which significantly reduces the workload of expert annotation. 

\subsubsection{Expert Annotation of Concepts} Given the pre-extracted concepts, we construct a group of expert annotators to refine, delete and add concepts to each exercise. The annotators are experienced teachers or teaching assistants from corresponding disciplines. Before the annotation setup, we present the extracted candidates as a reference for the most fine-grained concepts. All the annotators are required to complete the following sub-tasks: (1) Given an exercise and its relevant content, select or write out the knowledge it examines in fine-grained concepts; (2) Given the exercises in a same chapter, select or write out the medium-level concept to integrate them; (3) Given the exercises and videos in a same course, refine or write out an appropriate high-level concept to summarize the skill of this course. These multi-level annotations of concepts can be utilized to build a concept graph with a clear hierarchy.

\textbf{Quality Control.} Each term of data is assigned to two annotators for double checking. When merging fine-grained concepts provided by two annotators, we adopt the following strategy for quality assurance. For the \emph{Selected} concepts from the given candidates, we preserve the concept only if the two provided results are perfectly matched. For the \emph{Written} concepts, we preserve two types of them: a) they are perfectly matched; b) their semantic similarity is over $0.8$ and has a long common sub-sequence. In practice, after the merging guided by the above standards, we supplement with an expert annotation to refine and confirm the final results. We also monitor quality by calculating that the average consistency of the fine-grained concept labeling data across all courses is $0.673$ ($>0.6$), which indicates that the quality of the results is credible.

\begin{figure}
    \centering
    \includegraphics[width=0.85\linewidth]{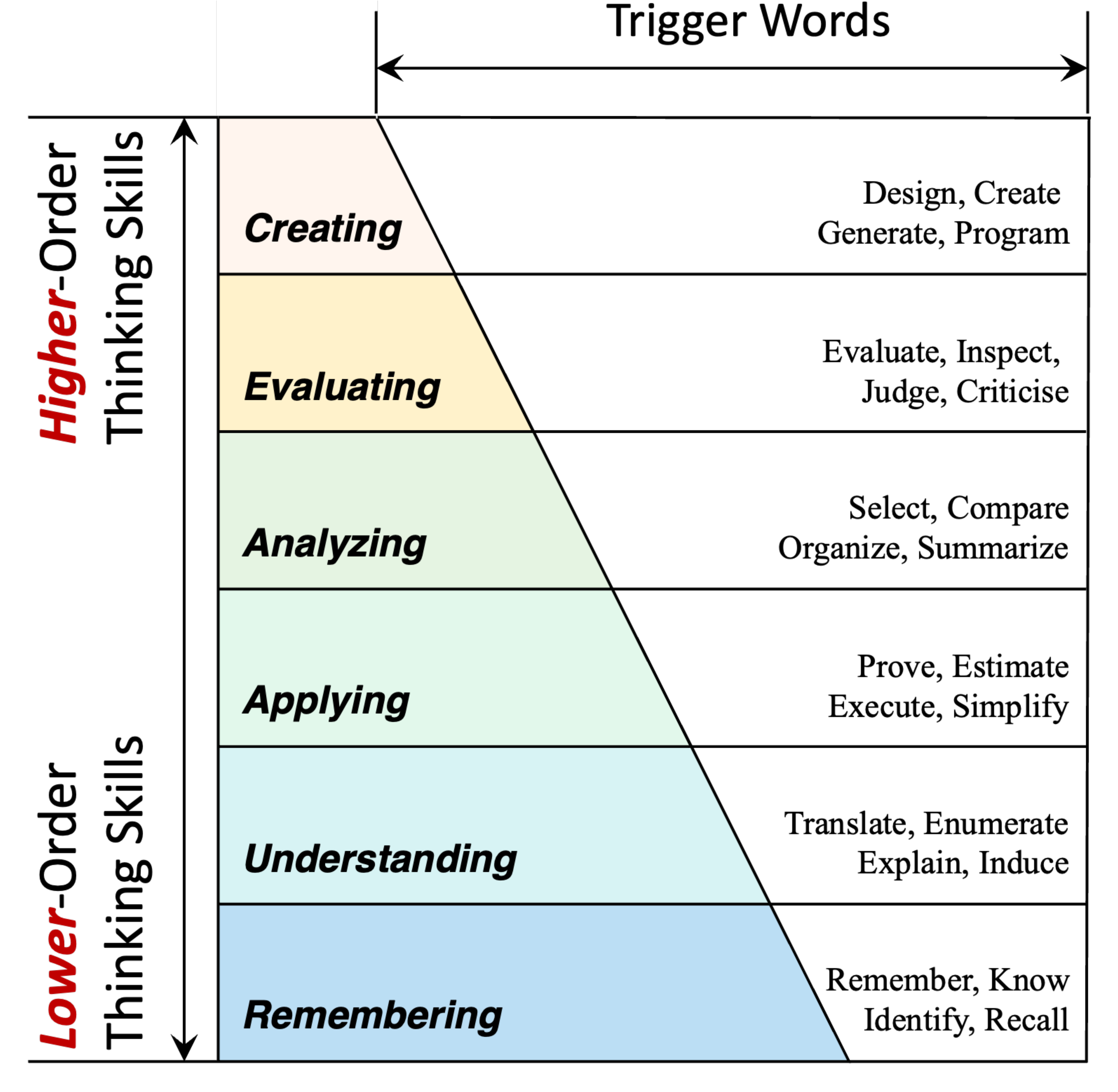}
    \caption{Bloom Cognitive Taxonomy. The trigger words are extracted via text mining and used as guidance in annotation.}
    \label{fig:bloom}
\end{figure}

\subsection{Cognitive Exercise Labeling}

The cognitive process of student learning is a rising feature in recent methods~\cite{shen2021learning,abdelrahman2022knowledge}, involving the difficulty of learning resources, students' cognitive patterns, etc. As there are no empirical attempts to label such features, we plan to take ingestion experience from widely accepted cognitive theories to provide a professional starting point for subsequent cognitive labeling. With the participation of experts in educational science (as co-authors of this paper), we decide to start with \emph{Bloom Cognitive Taxonomy}~\cite{krathwohl2002revision}, one of the most fundamental theories of modern educational cognition, as our labeling reference for subsequent annotations.

\subsubsection{Bloom Cognitive Taxonomy Annotation}

In \emph{Bloom Cognitive Taxonomy} (which was refined in 2001), a student's learning can be divided into six progressive levels as shown in Figure \ref{fig:bloom}: (1) \emph{Remembering.} The retention of specific, discrete pieces of information like facts and definitions or methodology. (2) \emph{Understanding.} The grasp of the meaning of a certain concept or instructional material. (3) \emph{Applying.} In this level, a student can use the information in a new (but similar) situation. (4) \emph{Analyzing.} A student can take apart the known and identify relationships. (5) \emph{Evaluating.} One can examine the information and make a judgment. (6) \emph{Creating.} Based on knowledge, one can create something new.

Despite the fact that the current assessment exercises are designed to estimate students' cognitive levels, it is still challenging to annotate a certain level of an exercise question without guidance. Therefore, we follow the suggestions from pedagogy and conduct text mining to determine a series of representative verbs for each cognitive level. Based on these trigger words, annotators are required to label each of exercise to the highest cognitive level.

\subsubsection{Quality Verification}

Each of the cognition-level labeling tasks is also assigned to two annotators. If there are inconsistencies in the merging process, we handle them in two categories: For the gaps of $1$ level or less, we take the larger of these values, as it partially examines the student's higher-order thinking. For the gaps of more than $2$ levels, we request a re-labeling and assign a third expert to make the determination. Finally, we also performed a consistency test. The consistency mean of the cognitive level labeling on all courses is $0.738$, which indicates that the quality of cognitive labeling is even better than concept annotation. After matching with the student data, the interception of a standard sequence of student behaviors in our data is shown in Table \ref{tab:dataset_behavior}.

\begin{table}[t]
\caption{The statistics of the diverse data in MoocRadar. \emph{Filter Rate} indicates the proportion of relevant content in raw data that is filtered out due to quality control.}
\label{tab:statistics}
\begin{tabular}{@{}l|l|r|c@{}}
\toprule
Item                      & Category   & Amount        & Filter Rate                 \\ \midrule
\multirow{2}{*}{Exercise} & Single     & 1,598      & 66.3\%                      \\
                          & Multi      & 915        & 58.2\%  \\ \hline
Student                   & -          & 14,224     & 91.6\% \\ \hline
\multirow{3}{*}{Concept}  & Coarse     & 120        & \multirow{3}{*}{-}          \\
                          & Middle     & 580        &                             \\
                          & Fine       & 5,600      &                             \\ \hline
Cognitive                   & -          & 2,513     & - \\ \hline
\multirow{2}{*}{Behavior} & Exercising & 1,752,319  & 95.0\%                      \\
                          & Learning   & 10,962,807 & 93.5\%                      \\ \bottomrule
\end{tabular}
\end{table}

\begin{table*}[ht]
\centering
\caption{The comparison of the existing educational datasets. \emph{Seq.Length}, \emph{Density} and \emph{Activity} correspond to the average length of students' exercising sequences, the quotient of the number of the datasets' concepts and exercises, and learning behaviors.}
\label{tab:comparison}
\begin{tabular}{@{}l|rrr|rrr|rr|c@{}}
\toprule
\multirow{2}{*}{Dataset} & \multicolumn{3}{c|}{Exercise-aspect Info.}                                                 & \multicolumn{3}{c|}{Concept-aspect Info.}                & \multicolumn{2}{c|}{Cognitive Info.}                       & \multirow{2}{*}{\begin{tabular}[c]{@{}c@{}}Manual\\ Annotation\end{tabular}} \\ \cmidrule(lr){2-9}
                         & \multicolumn{1}{r}{Seq.Length} & \multicolumn{1}{r}{Content} & \multicolumn{1}{r}{Structure} & \multicolumn{1}{|r}{Amount} & \multicolumn{1}{r}{Density} & \multicolumn{1}{r}{Graph} & \multicolumn{1}{|r}{Level} & \multicolumn{1}{r|}{Activity} &                                                                              \\ \midrule
ASSISTments2009          &             82.2                &      \xmark                     &                \xmark               &            123       &     0.004    &              \xmark             &         \xmark                  &        \xmark                            &        Yes                                                                      \\
ASSISTments2012             &          131.2                   &      \xmark                     &                \xmark               &            265    &    0.001        &              \xmark             &         \xmark                  &        \xmark                            &        Yes                                                                           \\ 
ASSISTments2015             &          35.4                   &      \xmark                     &                \xmark               &            \xmark    &      \xmark      &              \xmark             &         \xmark                  &        \xmark                            &        No                                                                           \\ \hline
Junyi Academy            &          104.6                  &       \cmark                     &          \cmark                     &            41     &     0.056      &               \xmark            &                \xmark           &         \xmark                           &             No                                                                 \\
Simulated-5              &           50.0                  &       \xmark               &          \xmark           &      5     &     0.100    &      \xmark         &    \xmark                  &        \xmark                   &    No                                                                   \\
Algebra 2005-2006       &           141.5                  &       \xmark               &          \xmark           &      523     &     0.103    &      \xmark         &    \xmark                  &        \xmark                   &    No                        \\ \hline
STATICS2011             &          107.8                  &       \cmark               &          \xmark           &      85    &     0.069     &      \xmark         &    \cmark                  &        \xmark                   &    No                                                                        \\
EdNet-KT                 &          121.5                   &     \cmark                       &              \cmark                 &      293         &      0.014       &    \xmark                       &          \xmark                 &     \cmark                               &        No                                                                      \\ \hline
MoocRadar                &          123.2     &        \cmark      &      \cmark                      &         5,600                      &        2.228                    &         \cmark                  &       \cmark                    &    \cmark                                &      Yes                                                                        \\ \bottomrule
\end{tabular}
\end{table*}

\subsection{Availability}

Researchers can get access to our repository on \url{https://github.com/THU-KEG/MOOC-Radar}. To facilitate the usage of data, we perform aggregation of multiple types of data and expose a series of tools with codes to simplify querying and supplementation. Specifically, there are three major components in the published repository:

$\bullet$ \textbf{Multiple Datasets}. As the exercising data can be associated with multi-level knowledge components (course, chapter, fine-grained concept) and other learning behaviors, we propose (1) the raw data with annotation, (2) cleaned exercising behavior records with different concept granularity, and (3) mixed student behavior of exercising and video watching, which can support knowledge tracing, learning recommendation, and other relevant tasks.

$\bullet$ \textbf{Easy-to-use Toolkit}. Meanwhile, we prepare a toolkit consisting of $12$ frequently used querying functions and APIs for convenient information retrieval and data structure construction. By combining multiple tools, developers can perform more complex queries and build more datasets according to their needs.

$\bullet$ \textbf{Reproduction Models}. We conduct several expert annotation stages and propose some models (such as the self-trained concept extraction). Therefore, we present the detailed annotation guidance and the source code of the model on the open page, which can support the reproduction and further extension of our repository.

We are continuing our work on the expert annotation for more disciplines and planning the introduction of more cognitive theories. Our repository will be continuously updated to follow these efforts. Moreover, we also conduct technical improvements of concept extraction tools as discussed in Section \ref{sec:concept_annotation}.

\begin{figure*}[htbp]
\centering
\subfigure[Student behavior length distribution.]{
\begin{minipage}[t]{0.31\textwidth}
\centering
\includegraphics[width=2.1in]{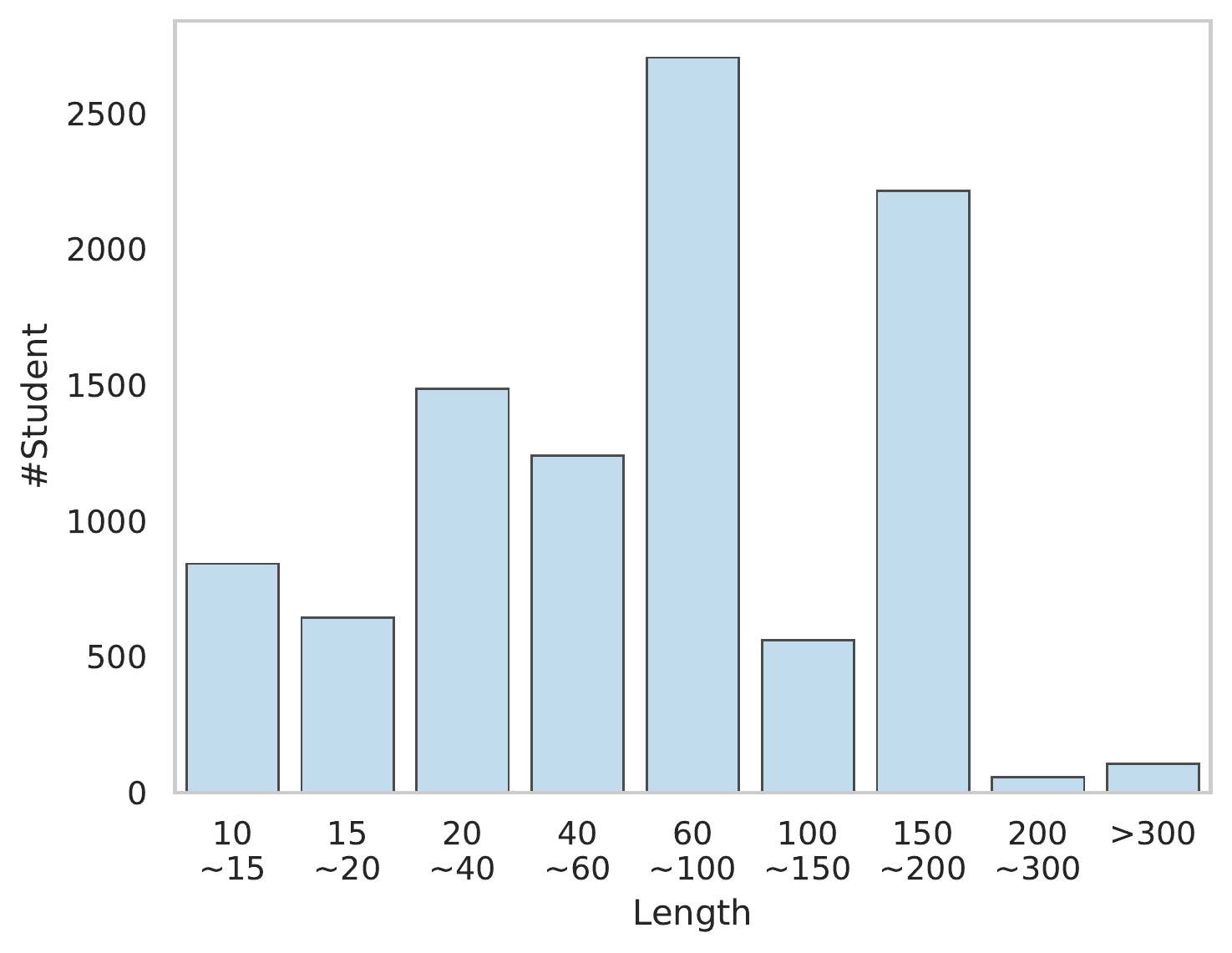}
\end{minipage}%
}%
\subfigure[Exercise accurate rate distribution.]{
\begin{minipage}[t]{0.31\textwidth}
\centering
\includegraphics[width=2.1in]{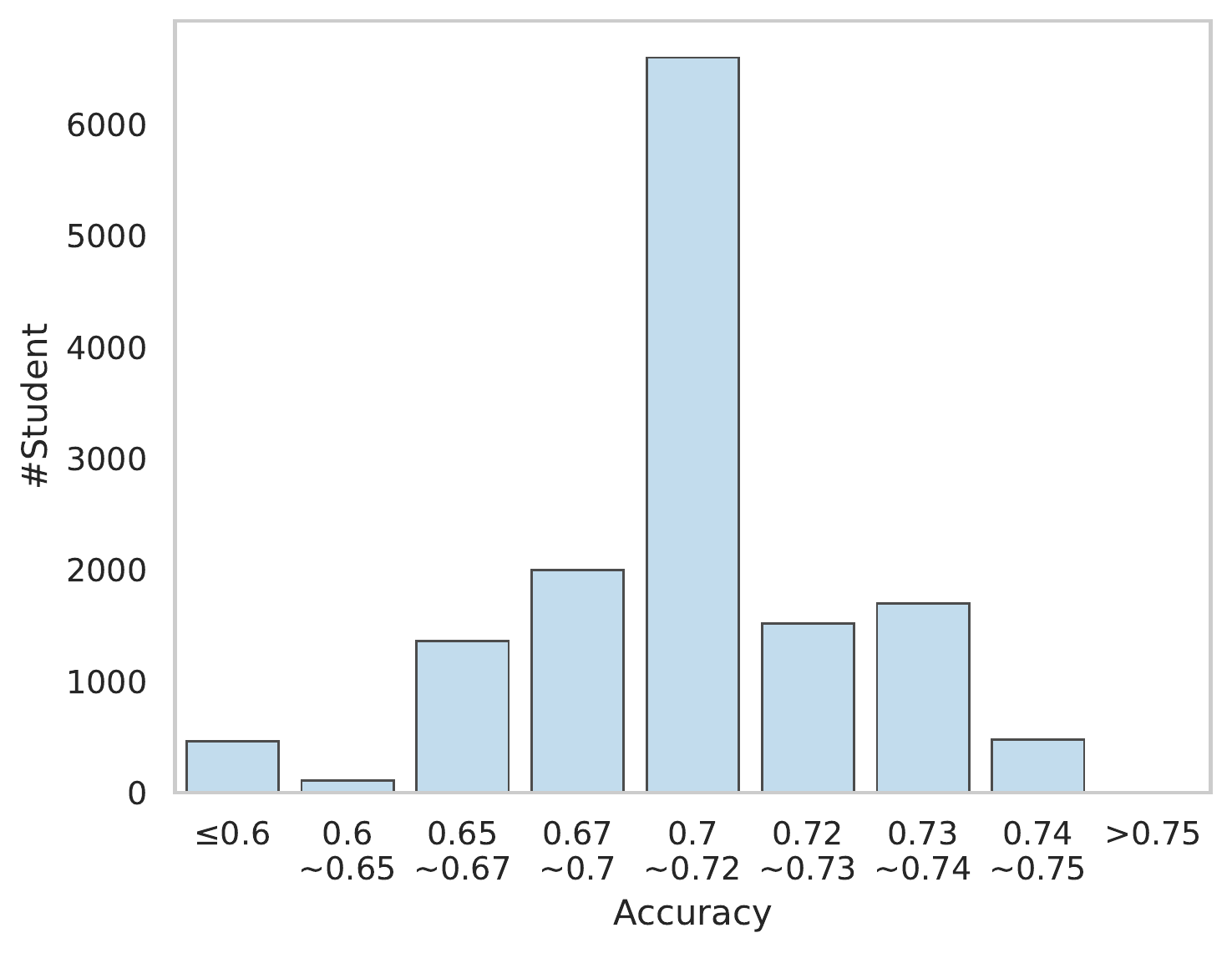}
\end{minipage}%
}%
\subfigure[Shared concept exercises distribution.]{
\begin{minipage}[t]{0.31\textwidth}
\centering
\includegraphics[width=2.1in]{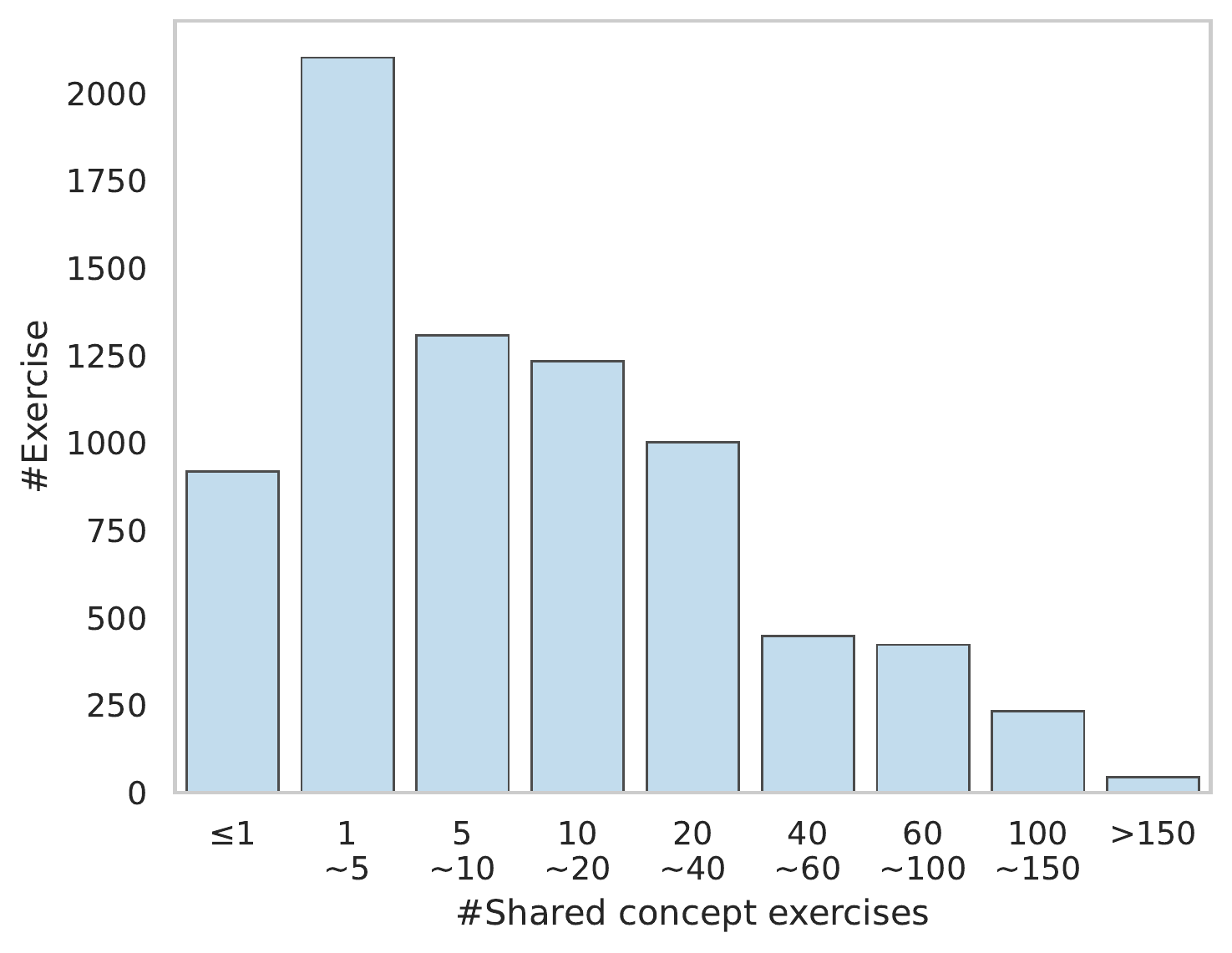}
\end{minipage}%
}%
\centering
\caption{The data distribution of exercises and the corresponding behaviors.}
\label{fig:behavior}
\end{figure*}

\section{Data Analysis}

In this section, we present the characteristics of our repository by analyzing its data statistics and distributions.

\textbf{Statistics.} Table \ref{tab:statistics} shows the detailed statistics of the diverse data in MoocRadar. Under strict data quality control, MoocRadar removes most of the unusable raw data and retains a sufficient amount of high-quality data (including multiple levels of knowledge) to support model training and evaluation. It's worth noting that behavior is the most filtered item, especially the exercising behaviors within it (with a $95.0\%$ filter rate).

\textbf{Comparison with other datasets.} We select several types of famous education datasets for comparison, including 1) ASSIST datasets (2009, 2012, 2015)~\cite{feng2009addressing,pardos2013affective}; 2) Public Challenges, \emph{i.e.}, Junyi Academy~\cite{JunyiOnlineLearningDataset}, Simulated-5~\cite{piech2015deep}, KDDcup Algebra 2005-2006~\cite{stamper2010challenge}; 3) Open Repositories, \emph{i.e.}, STATIC2011~\cite{koedinger2010data}, EdNET-KT~\cite{choi2020ednet}. 
We do not list the data source MOOCCubeX~\cite{yu2021mooccubex} in this comparison. 

We compare the various data sets respectively in terms of exercise, concept, and cognition in Table \ref{tab:comparison}. Overall, MoocRadar preserves the best data coverage of beneficial information, especially the most extensive knowledge concepts. Furthermore, we calculate two statistical features to present the usability of our repository. For the average length of students' exercising behaviors (\emph{Seq.Length}), MoocRadar is competitive with most of the current datasets. For the concept density of the exercises (\emph{Density}), MoocRadar has significant advantages over other existing datasets, \emph{i.e.}, each exercise corresponds to an average of 2.228 new concepts. Meanwhile, although many of the existing datasets have a large amount of raw data, they do require improvement in terms of information provision, concept granularity, etc., for supporting explorations.

\textbf{Exercising Behaviors and Concepts.} We also present the distributions of exercises and the corresponding behaviors in Figure \ref{fig:behavior}. From the observation, we can infer three features of our repository. First, most of the students have more than $60$ in exercising behaviors in our repository. According to the length of the behavioral sequence, students can be divided into two main groups, \emph{i.e.}, $60-100$ and $150-200$, and there is even a small group of students that complete over $300$ exercises, which again demonstrates the richness of the learning behavior in the dataset. Second, the correct rate is appropriately controlled.
After the filtering of student behaviors, the correctness of students' exercising is kept between $0.6$ and $0.75$, following a normal distribution. This indicates that cheating and other anomalies are effectively controlled, which mitigates the overfitting of existing models. Third, we calculate each exercise's relevant exercises according to the concept sharing, as shown in Figure \ref{fig:behavior} (c). Most of the exercises are linked by concepts to $2-20$ other exercises, thus building a graph of exercises with an appropriate density.
In general, MoocRadar's exercise data is of high quality and appropriately distributed.

\begin{figure}
    \centering
    \includegraphics[width=0.8\linewidth]{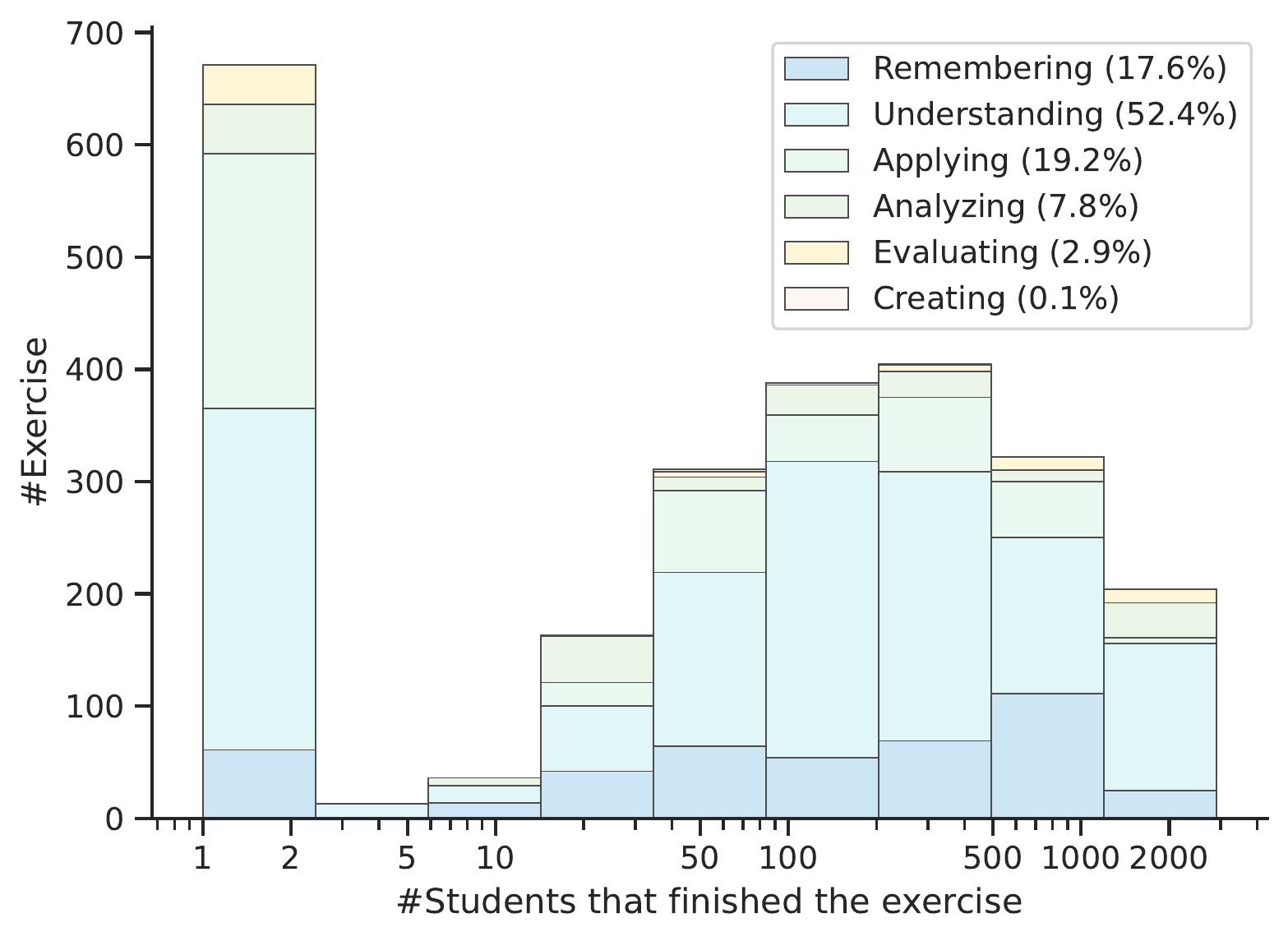}
    \caption{Stacked distribution of six cognitive levels' exercising behaviors of students.}
    \label{fig:cognitive}
\end{figure}

\textbf{Cognitive Level Distribution.} We also present the students' behavior over the six cognitive levels in Figure \ref{fig:cognitive}. From the statistics, we observe that most of the current student behaviors remain at a lower level (\emph{e.g.}, \emph{Remembering} and \emph{Understanding}), occupying $70.0\%$ of the total amount. Except for the remembering level, the behavior amount decreases as the cognitive level rises, following the common assumption from pedagogy~\cite{paramythis2003adaptive}. Meanwhile, there are still many exercises that have only been completed by some students, while the remaining exercises have been completed by a normal distribution of students. The highest-order creating accounts for only $0.1\%$ of the total, which may be due to the fact that the available data is mainly multiple-choice. In subsequent updates, we should add subjective questions to enrich the data types.



\section{Experiment}

In this section, we conduct primary experiments of both Knowledge Tracing and Cognitive Diagnosis to investigate the potential usage directions of the introduced features on our repository.

\subsection{Experimental Settings}

\subsubsection{Dataset}

We select the datasets with different concept granularity as the central experiment data basis. Meanwhile, we select the coarse-grained datasets with cognitive labels and video-watching behaviors as side information to further explore the effectiveness. In general, the datasets involve multi-level concepts (Coarse: $120$, Middle: $580$, Fine: $5600$), $2,513$ exercises, and $1,752,319$ behaviors. All the $6$ types of cognitive labels and over $6,567$ videos are utilized as side information when improving the methods.

\subsubsection{Baselines}

We produce several representative baselines on our datasets to observe primary results with the help of two public repositories: EduKTM\footnote{\url{https://github.com/bigdata-ustc/EduKTM}}~\cite{liu2021survey} and EduCDM\footnote{\url{https://github.com/bigdata-ustc/EduCDM}}~\cite{,wang2020neural,tong2021item}. 

For \emph{Knowledge Tracing} (KT) setting, we select (1) \textbf{DKT}~\cite{piech2015deep}: A RNN-based early method that introduce neural networks into this task. (2) \textbf{DKT+}~\cite{LS2018_Yeung_DKTP}: A refined KT method that considers the performance consistency. (3) \textbf{DKVMN}~\cite{zhang2017dynamic}: This method utilize a dynamic memory network to model students' behavior.

For \emph{Cognitive Diagnosis} (CD) setting, we select (1) \textbf{MIRT}~\cite{chalmers2012mirt}: A multi-dimension IRT method that can be applied in more environment. (2) \textbf{GDIRT}~\cite{embretson2013item}: A refined traditional IRT method that employs gradient descent. Note that the prior two are based on conventional linear regression. (3) \textbf{NCDM}~\cite{wang2020neural}: A famous adaption of neural networks in the cognitive diagnosis task.

\subsection{Result Analysis}

\begin{table}[t]
\small
\centering
\caption{Results of knowledge tracing (KT) and cognitive diagnosis (CD) models with different concept granularity. ``/'' means the performance is not applicable under this setting.}
\label{tab:overall}
\begin{tabular}{@{}l|l|l|l|l@{}}
\toprule
Setting             & Method                 &  Concept & AUC      & ACC      \\ \midrule
\multirow{9}{*}{KT} & \multirow{3}{*}{DKT}   & Coarse      & 0.815 &  0.831        \\
                    &                        & Middle      & 0.848 (+0.033) & 0.851 (+0.020)        \\
                    &                        & Fine        & /        & /        \\ \cline{2-5}
                    & \multirow{3}{*}{DKT+}  & Coarse      & 0.727 & 0.790 \\
                    &                        & Middle      & 0.741 (+0.014) & 0.819 (+0.029) \\
                    &                        & Fine        & 0.842 (+0.075) & 0.844 (+0.054) \\ \cline{2-5}
                    & \multirow{3}{*}{DKVMN} & Coarse      & 0.836  & 0.835  \\
                    &                        & Middle      & 0.863 (+0.027)  & 0.850 (+0.006)  \\
                    &                        & Fine        & 0.876 (+0.040)  & 0.862 (+0.027)  \\ \hline \hline
\multirow{9}{*}{CD} & \multirow{3}{*}{MIRT}  & Coarse      & 0.849 & 0.853 \\
                    &                        & Middle      & /        & /        \\
                    &                        & Fine        & /        & /        \\ \cline{2-5}
                    & \multirow{3}{*}{GDIRT}  & Coarse      &                      0.799 & 0.789 \\
                    &                        & Middle      &  0.832 (+0.033)       & 0.836  (+0.047)       \\
                    &                        & Fine        &  0.882 (+0.083)        &  0.867 (+0.078)        \\ \cline{2-5}
                    
                    & \multirow{3}{*}{NCDM}  & Coarse      & 0.816 & 0.837 \\
                    &                        & Middle      & 0.871 (+0.055) & 0.853 (+0.016) \\
                    &                        & Fine        &  0.885 (+0.069) & 0.868 (+0.029) \\\bottomrule
\end{tabular}
\end{table}

\begin{figure*}[htbp]
\centering
\subfigure[DKVMN with Cognitive features.]{
\begin{minipage}[t]{0.24\textwidth}
\centering
\includegraphics[width=1.7in]{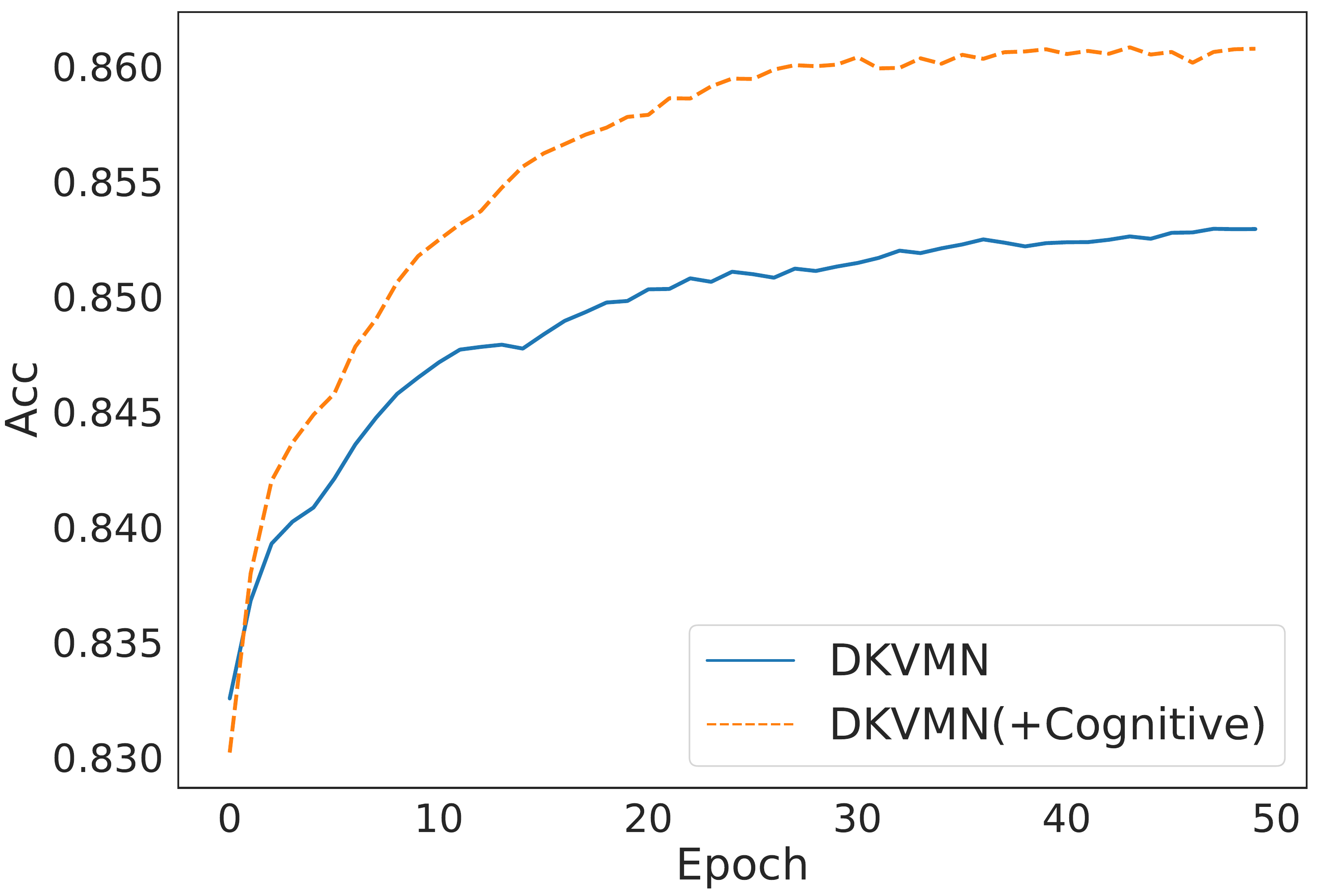}
\end{minipage}%
}%
\subfigure[NCDM with Cognitive features.]{
\begin{minipage}[t]{0.24\textwidth}
\centering
\includegraphics[width=1.7in]{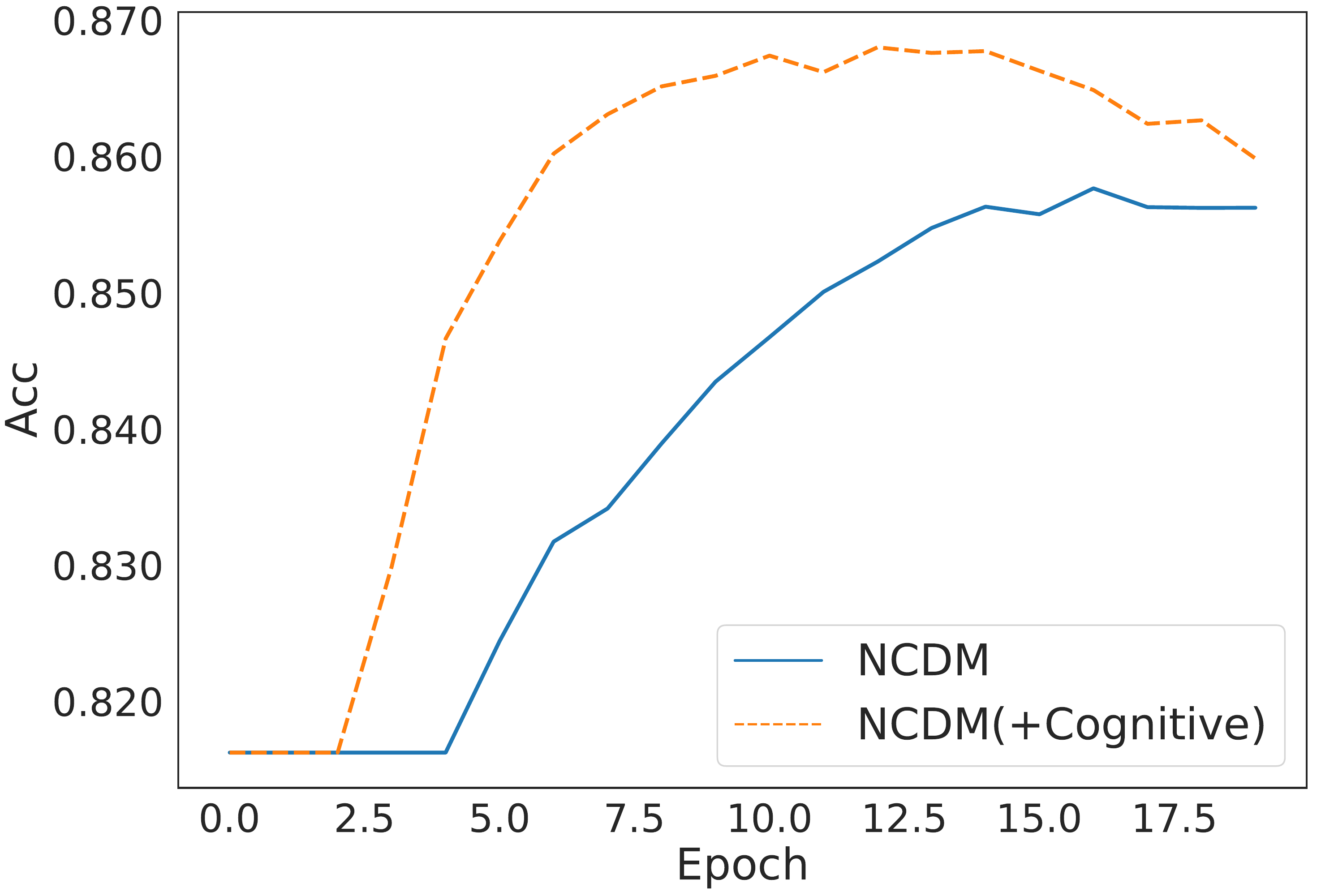}
\end{minipage}%
}%
\subfigure[DKVMN with Learning features.]{
\begin{minipage}[t]{0.24\textwidth}
\centering
\includegraphics[width=1.7in]{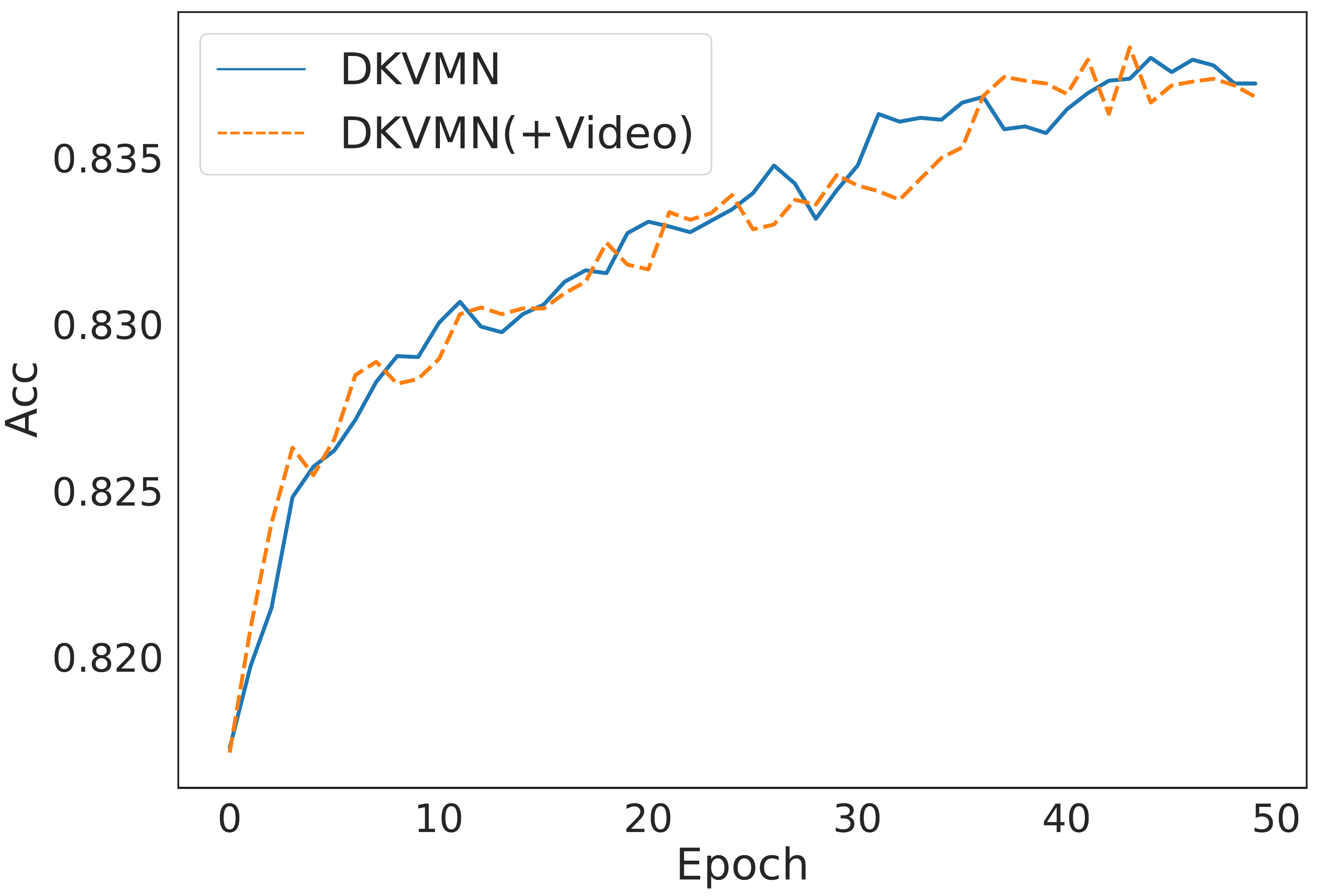}
\end{minipage}%
}%
\subfigure[NCDM with Learning features.]{
\begin{minipage}[t]{0.24\textwidth}
\centering
\includegraphics[width=1.7in]{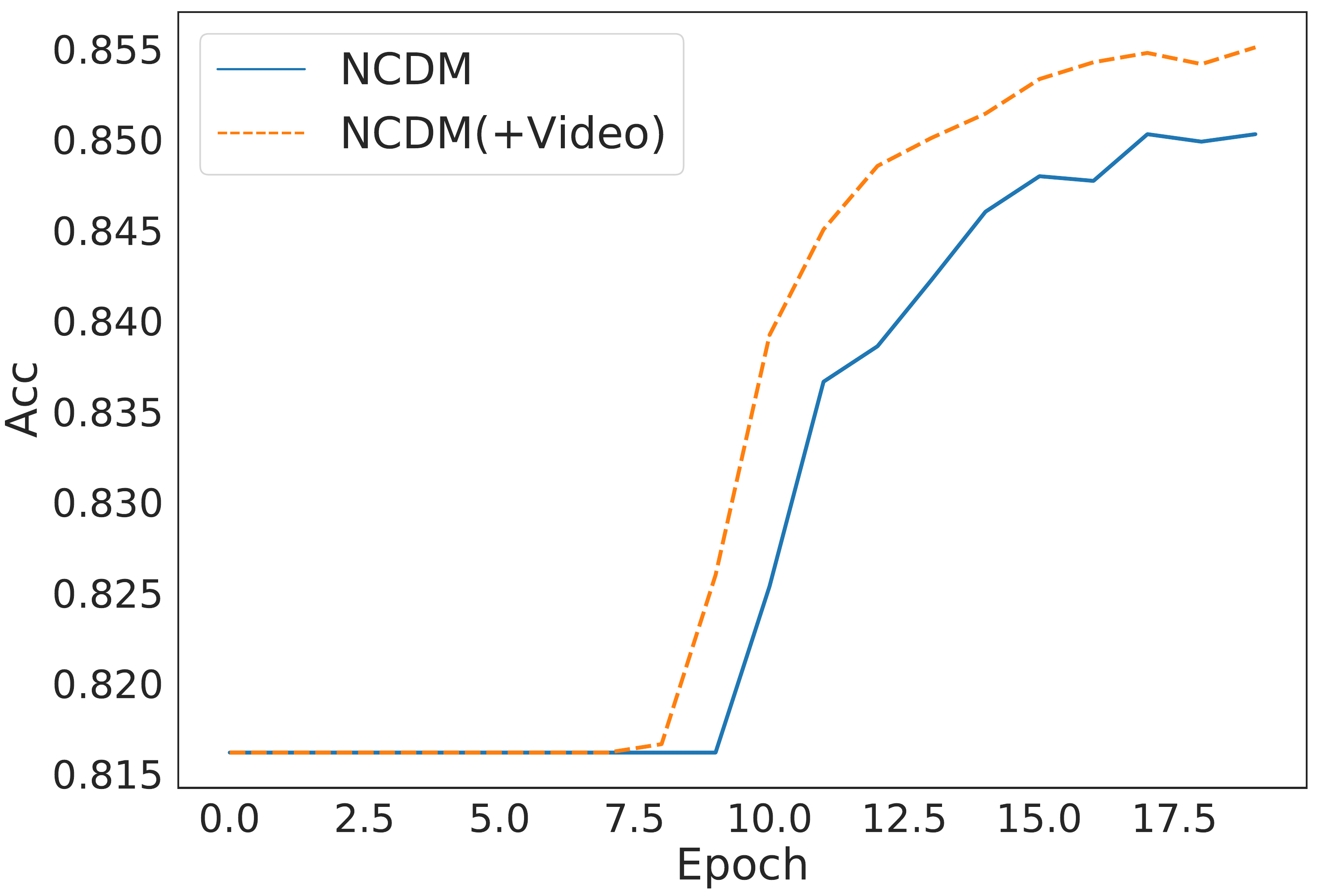}
\end{minipage}%
}%
\centering
\vspace{-0.112in}
\caption{The improved performance curve over training epochs of DKVMN and NCDM models.}
\label{fig:improve}
\end{figure*}

\subsubsection{Main Result}

The overall performance is shown in Table \ref{tab:overall}. Based on these results, we can infer some characteristics of our repository and discuss several major observations that may indicate the future directions of this topic.

First, MoocRadar can support diverse models of different topics and architectures (RNN, Regression, Transformer, Memory Network), which presents the convenience and usability of our repository. Meanwhile, as MoocRadar provides multi-level concepts, researchers can estimate their models' performance from more aspects and conduct more types of investigations.

Second, the performance of all models is improved when finer-grained concepts are used, which indicates the effectiveness of such annotation. One possible reason is that fine-grained concepts help the model better understand and explain student performance, while coarse-grained labeling sometimes confuses the model (students' inconsistent performance under the same coarse concepts is actually due to entirely different fine-grained knowledge).

Third, some of the conventional methods (\emph{e.g.}, DKT, MIRT) are not adaptable to fine-grained concept settings (due to excessive time and memory overhead), suggesting that recent attempts in the AI domain are essential. However, it is worth noting that the traditional method can still achieve quite competitive performance (\emph{e.g.}, GDIRT), which indicates that future research can continue to consider the combination of "traditional cognitive theory" and "advanced AI models" paradigm to advance the field.

\subsubsection{Model Improvement} 

Except for the fine-grained concepts, we conduct some attempts to improve existing models with cognitive labels and other learning behaviors. For the cognitive levels, we build a one-hot embedding and combine this as a new feature in the embedding layer to adjust the model. For the learning behaviors, we select the most recent five behaviors to form a sequence and then employ a learning behavior pre-training model~\cite{zhong2022towards} to obtain the embedding of it for extending the existing models.

Figure \ref{fig:improve} presents the performance (in terms of accuracy) of the improvement of DKVMN and NCDM, which are the top-performing models in the main experiments. We can observe that both of the two models are improved with the help of such side information, which can be summarized as: (1) The cognitive labels can effectively improve the models, which cover the whole process from the beginning to the convergence of training. Meanwhile, in addition to numerical boosts, adding this type of information can accelerate the convergence of the model, and it is also beneficial to the model's training stability. 
(2) The learning activities contribute to some of the models in a positive way after plausible modeling. Compared with cognitive labels, these data with raw noise do not enhance the stability of the model but still perform a lifting for learning-aware models (\emph{e.g.}, NCDM considers the learning difficulty). 
(3) The preliminary experimental results are based on the automatically calculated \emph{Accuracy}, which is a rather basic and general metric in all machine learning tasks. 

\section{Impact}

For the \emph{researchers in AI-driven Education}, MoocRadar provides a solid experimental basis to support the explorations of advanced models and methods. Recently, researchers have focused on many directions for improvement (\emph{e.g.}, concept relationships~\cite{chen2018prerequisite}, exercise structures~\cite{pandey2020rkt}, cognitive processes~\cite{abdelrahman2022knowledge}), and MoocRadar can effectively facilitate the implementation of these ideas, especially for those research processes previously limited by data.
Meanwhile, MoocRadar extends the existing task setting by introducing the cognitive levels, which may inspire researchers to introduce more cognitive science theories and enrich the student modeling task. 

For the \emph{developers of Intelligent Education Applications}, MoocRadar can play a more positive role and provides analytical insights for building more attractive educational products. The fine-grained concepts and cognitive levels in MoocRadar can be utilized as golden labels when a platform plans to conduct similar annotation on its own data~\cite{yu2022xdai}.
Furthermore, MoocRadar can serve as a fair benchmark for evaluating the pre-develop models of knowledge tracing, cognitive diagnosis, and learning recommendation.

Furthermore, MoocRadar is also a repository for researchers from \emph{Pedagogy} and \emph{Education Science} to conduct detailed analyses on online learning. In the mainstream trend of convergence between online and offline education~\cite{palvia2018online}, we hope MoocRadar can bring the experience of how to build an open and high-quality data bridge of AI techniques and educational research.

\section{Conclusion}

In this paper, we present MoocRadar, a fine-grained and multi-aspect knowledge repository that consists of $2,513$ exercises, $5,600$ concepts, and $14,224$ students' $12,715,126$ behavioral records for improving cognitive student modeling in MOOCs. Specifically, we host a construction framework and a series of standards for supporting the abundant expert annotation of fine-grained concepts and cognitive levels. We conduct statistical and experimental investigations on MoocRadar. The results show that our repository contains a high coverage of beneficial information and an appropriate distribution of data, while the provided rich features can indeed improve the current models with different architectures. The repository is now publicly available with a toolkit, which offers convenient access to our data for relevant researchers and developers.


\subsection*{Acknowledgement} 
This research project is supported by a grant from the Institute for Guo Qiang, Tsinghua University (2019GQB0003). Jie Tang is supported by NSFC distinguised young scholars (61825602). This research is also supported by National Natural Science Foundation of China (Grant No.62276154) and (Grant No.62277034).

\bibliographystyle{ACM-Reference-Format}
\bibliography{reference}

\end{document}